\documentclass[a4paper,10pt]{article}
\usepackage[T1]{fontenc}
\usepackage{parskip}
\usepackage[font={small}]{caption}
\usepackage[hidelinks]{hyperref}
\usepackage[margin=2.5cm]{geometry}
\usepackage{times}
\usepackage{amsmath,amssymb,amsthm}
\usepackage[affil-it]{authblk}
\usepackage{listings}
\usepackage[square,numbers]{natbib}

\usepackage{graphicx}      
\usepackage{natbib}        
\usepackage{amsmath,amsfonts,amssymb,bm} 
\usepackage{color}
\usepackage{algorithm}
\usepackage{algorithmic}
\usepackage{etoolbox}
\usepackage{siunitx}

\newcommand{\yAll}{\mathbf{y}}
\newcommand{\yAllExpand}{\left\{y_1, \dots, y_T\right\}}
\newcommand{\xAllExpand}{\left\{x_1, \dots, x_{T+1}\right\}}
\newcommand{\xAll}{\mathbf{x}}
\newcommand{\xAllTheta}{\tau}

\newcommand{\Transp}{\mathsf{T}}

\hypersetup{
	colorlinks=true,       
	linkcolor=blue,          
	citecolor=blue,        
	urlcolor=blue           
}

\date{\today}

\title{Variational State and Parameter Estimation\thanks{This work has been submitted to IFAC for possible publication.}} 

\author[1]{Jarrad Courts\thanks{\url{jarrad.courts@uon.edu.au}}}
\author[1]{Johannes Hendriks\thanks{\url{johannes.hendriks@newcastle.edu.au}}}
\author[1]{Adrian Wills\thanks{\url{Adrian.Wills@newcastle.edu.au}}}
\author[2]{Thomas~B.~Sch\"on\thanks{\url{thomas.schon@it.uu.se}},\thanks{This research was financially supported by the project \emph{Learning flexible models for nonlinear dynamics} (contract number: 2017-03807), funded by the Swedish Research Council and by \emph{Kjell och M\"arta Beijer Foundation} and by the project \emph{AI4Research} at Uppsala University.}}
\author[1]{Brett Ninness\thanks{\url{brett.ninness@newcastle.edu.au}}}

\affil[1]{University of Newcastle, School of Engineering, Australia}        
\affil[2]{Department of Information Technology, Uppsala University, Uppsala, Sweden}

\begin{document}
	
	\maketitle
		
	\begin{abstract}                
	 This paper considers the problem of computing Bayesian estimates of both states and model parameters for nonlinear state-space models. Generally, this problem does not have a tractable solution and approximations must be utilised. In this work, a variational approach is used to provide an assumed density which approximates the desired, intractable, distribution. The approach is deterministic and results in an optimisation problem of a standard form. Due to the parametrisation of the assumed density selected first- and second-order derivatives are readily available which allows for efficient solutions. The proposed method is compared against state-of-the-art Hamiltonian Monte Carlo in two numerical examples.
	\end{abstract}

\section{Introduction}
State-space models are a widely utilised and flexible class of models considered in many scientific and engineering applications \citep{Ljung1999}. Generally, identification of the parameters of state-space models remains a challenging task \citep{Ljung2010,Ninness2009}. In this paper, nonlinear state-space models of the form
\begin{subequations}
 \begin{align}
  x_{k+1} &\sim p(x_{k+1} \mid x_k, \theta),\\
  y_k &\sim p(y_k \mid x_k, \theta),
 \end{align}
\end{subequations}
are considered, where \( x_k \in \mathcal{R}^{n_x}\) and \( y_k \in \mathcal{R}^{n_y}\) are the state and measurement at time step \(k\), respectively, and \( \theta \in \mathcal{R}^{n_\theta}\) are model parameters. For clarity the presence of an input \( u_k \in \mathcal{R}^{n_u}\) is not shown, it is, however, allowed.

In this paper, given a sequence of measurements, \(\yAll \triangleq y_{1:T} \triangleq \yAllExpand\), the estimation of the state and model parameters is considered. Using a Bayesian approach, this amounts to obtaining the distribution \(p(\xAll, \theta \mid \yAll)\), given by
\begin{align} \label{eq:true_state_param_joint}
   p(\xAll, \theta \mid \yAll) = \frac{p(\yAll \mid \xAll, \theta) p(\xAll \mid \theta)p(\theta)}{p(\yAll)},
\end{align}
where \(\xAll \triangleq x_{1:T+1} \triangleq \xAllExpand\), and \( p(\theta)\) is a prior distribution over the parameters. 

From \(p(\xAll, \theta \mid \yAll)\), distributions of both parameters and states can be obtained via marginalisation. For example, the marginal distribution \(p(\theta \mid \yAll) \) is found via
\begin{align} \label{eq:true_param_dist_via_marginlisation}
  p(\theta \mid \yAll) = \int p(\xAll, \theta \mid \yAll) d\xAll.
\end{align}

The difficulty with both \eqref{eq:true_state_param_joint} and \eqref{eq:true_param_dist_via_marginlisation} is that they are generally intractable. This is due to both the arbitrary form of the distributions considered and the nonlinear integrals involved. As such, these distributions must be approximated. Many differing approximations exist, one class of approaches is to utilise particle-based Monte Carlo methods, an example of this is Hamiltonian Monte Carlo (HMC) \citep{Neal2011,Hendriks2020} methods.

In this paper, an alternative approach, known as variational inference (VI) \citep{Jordan1999,Blei2017}, to approximate \(p(\xAll, \theta \mid \yAll)\) is followed. The contribution of this paper is to use VI to approximate \(p(\xAll, \theta \mid \yAll)\). Using a Gaussian assumed density, and a suitable approximation for the integrals, this provides a deterministic method of approximating \(p(\xAll, \theta \mid \yAll)\) that can be efficiently performed using gradient-based optimisation with exact first- and second-order derivatives.

\section{Variational Inference}\label{sec:VI}
Variational inference is a widely used method to approximate, potentially intractable, posterior distributions with a parametric density of an assumed form. The numeric values of this parametrisation are obtained by maximising a likelihood lower bound \citep{Blei2017}. In this section, the use of VI in approximating \( p(\xAll, \theta \mid \yAll) \) is examined.

As \( p(\xAll, \theta \mid \yAll) \) is intractable, it is approximated using an assumed density, parameterised by \(\beta\), and denoted as \(q_\beta(\xAll, \theta)\). The quality of this approximation is measured by
\begin{align}
 \text{KL}[ q_\beta(\xAll, \theta)  \mid\mid p(\xAll, \theta \mid \yAll )],
\end{align}
the Kullback-Leiber (KL) \citep{Kullback1951} divergence of \( p(\xAll, \theta \mid \yAll )\) from \( q_\beta(\xAll, \theta) \). As it is desired that  \( q_\beta(\xAll, \theta) \) is close to \( p(\xAll, \theta \mid \yAll )\) the aim is to find a \(\beta\), denoted \(\beta^\star\), which minimises this KL divergence and is given by
\begin{align}
 \beta^\star = \arg \min_{\beta}  \quad \text{KL}[ q_\beta(\xAll, \theta)  \mid\mid p(\xAll, \theta \mid \yAll )].
\end{align}

However, as \( p(\xAll, \theta \mid \yAll )\) is intractable \(\text{KL}[ q_\beta(\xAll)  \mid\mid p(\xAll, \theta \mid \yAll )]\) cannot be directly minimised or even evaluated. Instead, the concept of VI is used to, equivalently, maximise a likelihood lower bound. 

Using conditional probability, this lower bound can be found by expressing the log-likelihood, \(\log p(\yAll)\), as
\begin{align} \label{eq:log_cond_prob}
 \log p(\yAll) &= \log p(\xAll, \theta, \yAll) - \log p(\xAll, \theta \mid \yAll ).
\end{align} 
Through addition and subtraction of \( \log q_\beta(\xAll, \theta) \) to the right-hand side of \eqref{eq:log_cond_prob} this leads to 
\begin{align} \label{eq:LL_sum}
 \log p(\yAll) &= \log \frac{ p(\xAll, \theta, \yAll)  }{ q_\beta(\xAll, \theta) } + \log \frac{ q_\beta(\xAll, \theta) }{ p(\xAll, \theta \mid \yAll )}.
\end{align}
As \(\log p(\yAll) \) is independent of \(\xAll\) and \(\theta\), the log-likelihood can alternatively be given by
\begin{align} \label{eq:LL_independence}
 \log p(\yAll) = \int  q_\beta(\xAll, \theta) \log p(\yAll) d \xAllTheta,
\end{align}
where \(\xAllTheta = [\xAll^\Transp, \theta^\Transp]^\Transp\).

Substituting \eqref{eq:LL_sum} into the right-hand side of \eqref{eq:LL_independence} arrives at
\begin{align}
 \log p(\yAll) = &\int q_\beta(\xAll, \theta) \log \frac{ p(\xAll, \theta, \yAll)  }{ q_\beta(\xAll, \theta) } d\xAllTheta  \notag\\
     &+ \int  q_\beta(\xAll, \theta) \log \frac{ q_\beta(\xAll, \theta) }{ p(\xAll, \theta \mid \yAll )} d\xAllTheta,
\end{align}
which will be expressed as 
\begin{align} \label{eq:ll_equal_L_plus_KL}
 \log p(\yAll) = \mathcal{L}\left(\beta\right) + \text{KL}[ q_\beta(\xAll, \theta)  \mid\mid p(\xAll, \theta \mid \yAll )],
\end{align}
where 
\begin{align}
 \mathcal{L}\left(\beta\right) = \int q_\beta(\xAll, \theta) \log \frac{ p(\xAll, \theta, \yAll)  }{ q_\beta(\xAll,\theta) } d\xAllTheta.
\end{align} 
From \eqref{eq:ll_equal_L_plus_KL}, and as any KL divergence is non-negative, \(\mathcal{L}\left(\beta\right)\) is a lower bound to the log-likelihood \(\log p(\yAll)\).

Furthermore, from \eqref{eq:ll_equal_L_plus_KL} we have
\begin{align}
 \text{KL}[ q_\beta(\xAll, \theta)  \mid\mid p(\xAll, \theta \mid \yAll )] =  \log p(\yAll) - \mathcal{L}\left(\beta\right),
\end{align}
and therefore
\begin{subequations}
 \begin{align}
  \beta^\star &= \arg \min_{\beta} \quad \text{KL}[ q_\beta(\xAll, \theta)  \mid\mid p(\xAll, \theta \mid \yAll )] \\
     &= \arg \min_{\beta} \quad \log p(\yAll) - \mathcal{L}\left(\beta\right) \\
     &= \arg \max_{\beta} \quad \mathcal{L}\left(\beta\right).     
 \end{align}
\end{subequations}
The important property of this is \(\mathcal{L}\left(\beta\right)\) does not involve the intractable distribution, \( p(\xAll, \theta \mid \yAll )\), rather only the assumed distribution \(q_\beta(\xAll, \theta)\) which is selected in a tractable form.

The desired approach is therefore to find a value for \(\beta\), that maximises \(\mathcal{L}\left(\beta\right) \) via
\begin{align} \label{eq:highest_level_optim}
 \beta^\star &= \arg\max_{\beta}  \quad \mathcal{L}(\beta),
\end{align}
for an assumed density form. This provides an approximation for \( p(\xAll, \theta \mid \yAll )\) given by \(q_\beta(\xAll, \theta)\) as desired.

However, as given, this approach is computationally intractable for any large quantity of time steps. This is because \( q_{\beta}(\xAll,\theta) \) is a full dense distribution, the dimension of which continues to grow. Fortunately, by expressing \(\mathcal{L}\left(\beta\right)\) as
 \begin{align}
  \mathcal{L}\left(\beta\right) &= \int q_\beta(\xAll, \theta) \log \frac{ p(\xAll, \theta, \yAll)  }{ q_\beta(\xAll,\theta) } d\xAllTheta \notag \\
  &= \int q_\beta(\xAll, \theta) \log p(\xAll, \theta, \yAll) d\xAllTheta \notag\\
  &\quad- \int q_\beta(\xAll, \theta) \log q_\beta(\xAll,\theta) d\xAllTheta,
 \end{align}
and utilising the Markovian nature of state-space models (see Appendix~\ref{sec:integral_decomposition} for details) to give 
 \begin{align}
  &\int q_\beta(\xAll, \theta) \log p(\xAll, \theta, \yAll) d\xAllTheta \notag\\
  &\quad=\int q_\beta(x_1, \theta) \log p(x_1, \theta) d[x_1, \theta] \notag \\
  &\quad\quad + \sum_{k=1}^{T}  \int q_\beta(x_{k:k+1}, \theta) \log p(x_{k+1}, y_k \mid x_k, \theta) d[x_{k:k+1}, \theta] \notag\\
  &\quad=I_1\left(\beta\right) + I_{23}\left(\beta\right),
 \end{align}
and 
 \begin{align}
  &\int q_\beta(\xAll, \theta) \log q_\beta(\xAll,\theta) d\xAllTheta \notag \\
  &\quad= \sum_{k=1}^{T}\int q_\beta(x_{k+1}, x_k, \theta) \log q_\beta(x_{k+1}, x_k, \theta) d[x_{k:k+1}, \theta] \notag\\
  &\quad\quad- \sum_{k=2}^{T}\int q_\beta( x_k, \theta) \log q_\beta(x_k, \theta) d[x_{k}, \theta]  \notag \\
  &\quad= I_4\left(\beta\right),
 \end{align}
we have
\begin{align}
  \mathcal{L}\left(\beta\right) = I_1\left(\beta\right) + I_{23}\left(\beta\right) - I_4\left(\beta\right),
\end{align}
without introducing any approximations.

Therefore, calculation of \(\mathcal{L}\left(\beta\right)\) does not require the full distribution, rather only each \(q_\beta(\theta, x_k, x_{k+1})\) and as such remains tractable to calculate for large quantities of time steps. 
 
The focus of this paper correspondingly shifts from the, intractable, aim of obtaining \(q_\beta(\xAll, \theta)\), to the tractable aim of obtaining each \(q_\beta(\theta, x_k, x_{k+1})\).

\section{Assumed Gaussian Distribution and Tractable Approximation} \label{sec:assumed_density_and_approx}
In this section, the parametric assumed density chosen and the subsequent formation of a standard form optimisation problem is examined. When selecting the parametric form of \(q_\beta(\theta, x_k, x_{k+1})\) there are two conflicting goals. Firstly, the parametrisation chosen should be sufficiently flexible to well represent \(p(\theta, x_k, x_{k+1})\). Secondly, the parametrisation should be chosen to allow the optimisation to be efficiently performed.
 
In this paper, we have chosen to use a multivariate Gaussian parameterised as
\begin{align}
 q_{\beta_k}\left(\theta,x_k,x_{k+1}\right) = \mathcal{N}\left(
 \begin{bmatrix}
  \theta \\
  x_k \\
  x_{k+1}
 \end{bmatrix} ; 
 \begin{bmatrix}
  \mu_k \\ 
  \bar{\mu}_k \\
  \tilde{\mu}_k
 \end{bmatrix},
 P_k^{\frac{\Transp}{2}} P_k^{\frac{1}{2}}
 \right),
\end{align}
where \( \mu_k \in \mathcal{R}^{n_\theta}\), \( \bar{\mu}_k \in \mathcal{R}^{n_x}\), \(\tilde{\mu}_k  \in \mathcal{R}^{n_x}\), \(P_k^{\frac{1}{2}} \in \mathcal{R}^{(n_\theta+2n_x) \times (n_\theta+2n_x)}\), and
\begin{align}
 P_k^{\frac{1}{2}} = \begin{bmatrix}
  A_k & B_k & C_k\\
  0 & D_k & E_k \\
  0 & 0 & F_k
 \end{bmatrix},
\end{align}
where \( A_k, D_k, F_k\) are upper triangular. The parameter \(\beta_k\) is therefore given as 
\begin{align}
 \beta_k =  \{ \mu_k, \bar{\mu}_k, \tilde{\mu}_k A_k, B_k, C_k, D_k, E_k, F_k \},
\end{align}
and related to \(\beta\) via
\begin{align}
 \beta = \{\beta_1, \beta_2, \dots, \beta_T \}.
\end{align}

Importantly, parameterising each joint distribution using the Cholesky factor, \(P_k^{\frac{1}{2}} \), allows the components of \({I}_{23}\left(\beta\right)\) corresponding to each time step to be easily approximated using Gaussian quadrature as
\begin{subequations} \label{eq:approx_integral}
 \begin{align} 
  \hat{I}_{23_k}\left(\beta\right) &= \sum_{j = 1}^{n_s} w_j  \log p(x_{k+1}^j, y_k \mid x_k^j, \theta^j_k),
 \end{align}
\end{subequations}
where $w_j$ is a weight, and the $n_s$ sigma points are denoted as \( \xi^j  \in \mathcal{R}^{n_\theta+2n_x}\) where
\(
\xi^j = \begin{bmatrix}
	(\theta^j_k)^\Transp, &
	(x_k^j)^\Transp,        &
	(x_{k+1}^j)^\Transp
\end{bmatrix}^\Transp
\)
is given by linear combinations of the joint mean 
\(
\begin{bmatrix}
 \mu_k^\Transp, 
 \bar{\mu}_k^\Transp, \tilde{\mu}_k^\Transp
\end{bmatrix}^\Transp
\)
and the columns of \(P_k^{\frac{\Transp}{2}}\). The sigma points, \(\xi^j\), being linear combinations of the elements of \(\beta_k\) is important as it significantly simplifies the calculation of both first- and second-order derivatives used in the optimisation.

Furthermore, due to the Gaussian assumption \(I_4\left(\beta\right)\), and its derivatives, are available in closed form. With a suitable prior \(I_1\left(\beta\right)\) is also available in closed form, alternatively it can be similarly approximated using Gaussian quadrature.
 
This allows \( \mathcal{L}\left(\beta\right)\) to be approximated by \( \hat{\mathcal{L}}\left(\beta\right) \) as
\begin{align}
 \hat{\mathcal{L}}\left(\beta\right) &= I_1\left(\beta\right) + \hat{I}_{23}\left(\beta\right) - I_4\left(\beta\right).
\end{align}
 
This parametrisation is, however, over-parameterised as the marginal distributions \(q_\beta(\theta_k)\), \(q_\beta(x_k)\) and \(q_\beta(\theta,x_k)\) can be calculated from either \( q_{\beta}\left(\theta_{k+1},x_{k-1},x_{k}\right) \) or \\
\(q_{\beta}\left(\theta_k, x_k,x_{k+1}\right) \). Hence, the constraints defined by the feasible set
 \begin{align}
	\bar{\Omega} := \{\beta \in \mathcal{R}^{n_\beta} \mid \bar{c}_k\left(\beta\right) = 0, \quad k = 1, \dots, T-1 \},   	
\end{align}   
where \(n_\beta = T\left(n_\theta+2n_x + \frac{\left(n_\theta+2n_x\right)\left(n_\theta+2n_x+1\right)}{2}\right)\) and
\small
\begin{align*}
	\bar{c}_k\left(\beta\right) = \begin{bmatrix}
		 \mu_{k+1} - \mu_{k+1} \\
		 \tilde{\mu}_{i+1} - \bar{\mu}_i \\
		 A_k^\Transp A_k - A_{k+1}^\Transp A_{k+1}\\
		 A_k^\Transp C_k - A_{k+1}^\Transp B_{k+1} \\
		 C_k^\Transp C_k + E_{k}^\Transp E_{k} + F_k^\Transp F_k - B_{k+1}^\Transp B_{k+1} - D_{k+1}^\Transp D_{k+1}
	\end{bmatrix},
\end{align*}
\normalsize
are required to ensure consistency. 

However, as each \(A_k\), being a Cholesky factor of a covariance matrix, is non-singular the constraints set can be simplified to
\begin{align}
	\Omega := \{\beta \in \mathcal{R}^{n_\beta} \mid c_k\left(\beta\right) = 0, \quad k = 1, \dots, T-1 \},	
\end{align} 
where
\small
\begin{align*}
	c_k\left(\beta\right) = \begin{bmatrix}
		\mu_{k+1} - \mu_{k+1} \\
		\tilde{\mu}_{i+1} - \bar{\mu}_i \\
		A_k - A_{k+1} \\
		C_k - B_{k+1} \\
		C_k^\Transp C_k + E_{k}^\Transp E_{k} + F_k^\Transp F_k - B_{k+1}^\Transp B_{k+1} - D_{k+1}^\Transp D_{k+1}
	\end{bmatrix}.
\end{align*}
\normalsize

Using this, an approximation of \eqref{eq:highest_level_optim} is given by
\begin{subequations} \label{eq:approx_optim}
	\begin{align} 
		\hat{\beta} = \arg\max_{\beta}  \quad &\hat{\mathcal{L}}(\beta), \\
	  	\text{s.t.} \quad &\beta \in \Omega.
	\end{align}
\end{subequations}
This constrained optimisation problem is of a standard form and can be efficiently solved using exact first- and second-order derivatives without more approximations to find a local maximum. The calculation of derivatives required can be performed using automatic differentiation \citep{Griewank2008} without manual effort. Due to the parametrisation chosen, this can be efficiently performed using standard tools. For the numerical examples provided CasADi \citep{Andersson2018} has been used. 

The result of this optimisation is numerical values for each
\begin{align}
 q_{\hat{\beta}_k}\left(\theta,x_k,x_{k+1}\right) = \mathcal{N}\left(
 \begin{bmatrix}
  \theta \\
  x_k \\
  x_{k+1}
 \end{bmatrix} ; 
 \begin{bmatrix}
  \mu_k \\ 
  \bar{\mu}_k \\
  \tilde{\mu}_k
 \end{bmatrix},
 P_k^{\frac{\Transp}{2}} P_k^{\frac{1}{2}}
 \right),
\end{align}
which approximates each \( p\left(\theta,x_k,x_{k+1} \mid \yAll \right) \). Due to the Gaussian form selected, marginal and conditional distributions of both the parameters and states can be readily extracted as desired.

The proposed approach is summarised in Algorithm~\ref{alg:general_ID}. 
\begin{algorithm}[t] 
 \caption{State and Parameter Estimation}
 \label{alg:general_ID}
 \begin{algorithmic} 
  \renewcommand{\algorithmicrequire}{\textbf{Input:}}
  \renewcommand{\algorithmicensure}{\textbf{Output:}}
  \REQUIRE Measurements \(y_{1:T}\), prior \(p(x_1,\theta)\), initial estimate of \(\beta\)
  \ENSURE Each \(q_{\hat{\beta}_k}\left(\theta,x_k,x_{k+1}\right)\) for \( k \in 1, \dots, T\)
  \STATE - Obtain \(\hat{\beta}\) from \eqref{eq:approx_optim} initialised at \(\beta\)
  \STATE - Extract each \(\hat{\beta}_k\) from \(\hat{\beta}\).
 \end{algorithmic}
\end{algorithm} 

\section{Numerical Examples} \label{sec:examples}
In this section, we provide two numerical examples of the proposed approach and compare the results against HMC. The HMC results have been calculating using STAN \citep{Carpenter2017} as detailed in \cite{Hendriks2020}.

\subsection{Stochastic Volatility}
In this example, estimation of \(\theta = [a, b, \log(\sqrt{c})]\) for the following stochastic volatility model:
\begin{subequations}
 \begin{align}
  x_{k+1} &= a + b x_k + \sqrt{c} w_k, \\
  y_k  &= \sqrt{e^{x_k}}v_k,
 \end{align}
\end{subequations}
where $w_k \sim \mathcal{N}(0,1)$ and $v_k \sim \mathcal{N}(0,1)$ is considered using 726 simulated measurements.

For the proposed method fifth-order cubature \citep{Jia2013} is used to perform the Gaussian quadrature approximations and the optimisation routine is run until convergence to a locally optimal point requiring \SI{17.6}{\second}. The HMC results are obtained using \num{8e3} iterations with half discarded at warm-up requiring \SI{87.0}{\second}.

\begin{figure}[!t]
 \centering
 \includegraphics{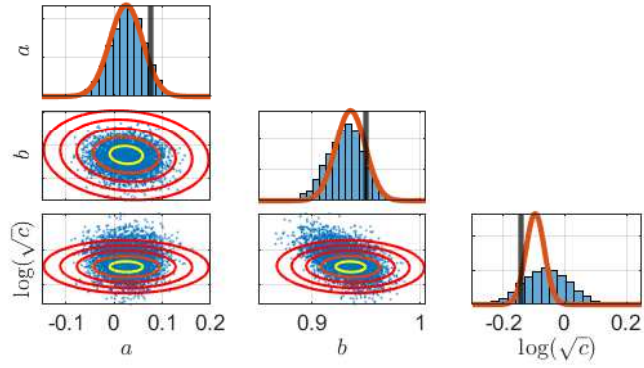}
 \caption{Distribution of \(\theta\) for both the proposed approach (contour plots) and HMC (scatter plots / histograms). The true parameters are shown as vertical lines.}
 \label{fig:bitcoin_param_dists}
\end{figure}

In Figure~\ref{fig:bitcoin_param_dists} the marginal distributions of \(\theta\) using both the proposed approach and HMC are shown. While the marginal distributions for parameters \(a\) and \(b\) closely match, the VI-based method is overconfident in the distribution of \(\log(\sqrt{c})\) and cannot closely match the non-Gaussian relationships between the parameters. 

\begin{figure}[!t]
 \centering
 \includegraphics{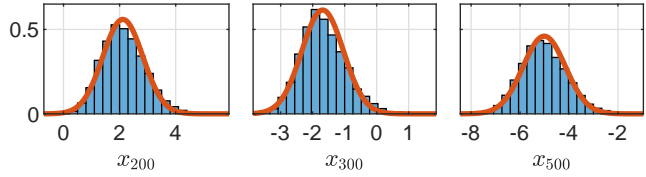}
 \caption{State distributions for three time steps using the proposed approach (curve) and HMC (histogram) illustrating the high level of similarity. }
 \label{fig:bitcoin_state_dists}
\end{figure}

Despite this the state distributions obtained using both methods closely match. Figure~\ref{fig:bitcoin_state_dists} highlights this for three differing time steps, which are representative of the results for other time steps.

In this example, the proposed method is demonstrated on a nonlinear and non-Gaussian example and is, generally, representative of the results obtained using HMC at a lower computational cost. This, however, comes with a trade-off, when the underlying distribution is non-Gaussian, the Gaussian assumption of the proposed method may produce overconfident results as shown for the distribution of \(\log(\sqrt{c})\).
 
\subsection{Inverted Pendulum}
In this section, the proposed method is applied to a rotational inverted pendulum, or Furata pendulum \citep{Furuta1992}, using real data collected from a Quanser QUBE—Servo 2. The estimation of a four-dimensional state vector, six parameters \(\phi = [J_r, J_p, K_m, R_m, D_p, D_r]^\Transp\), and a seven dimensional covariance matrix \(\Pi\) associated with coupled additive Gaussian noise of the process and measurement models is considered using 375 measurements. For further details of the model refer to Appendix~\ref{sec:pendulum_model}.

For the proposed approach, a third order unscented-transform \citep{Julier1997,Wan2000} with \(\alpha = 1\), \(\kappa = 0\), and \(\beta = 0\) is used to perform the Gaussian quadrature required. A point estimate for \(\Pi\) is also considered, this highlights the flexibility of the proposed approach which allows for distributions of subsets of parameters alongside point estimates of others. In this example, we also exploit the additive form of the state-space model, this allows for a Hessian to be efficiently approximated using only Jacobian’s. Identification problems for systems with additive Gaussian noise commonly use this approach \citep{Kokkala2015,Saerkkae2013}. While the calculations for this example are not fully detailed in Section~\ref{sec:assumed_density_and_approx}, they represent minor modifications of the resultant optimisation problem.

For the HMC results, \num{10e3} iterations are used to obtain distributions for all states and parameters. As the HMC results calculate a distribution for each element of \(\Pi\), while the proposed method considers a point estimate, it is expected the results of the proposed method and HMC slightly differ.

\begin{figure*}[!t]
 \centering
  \includegraphics{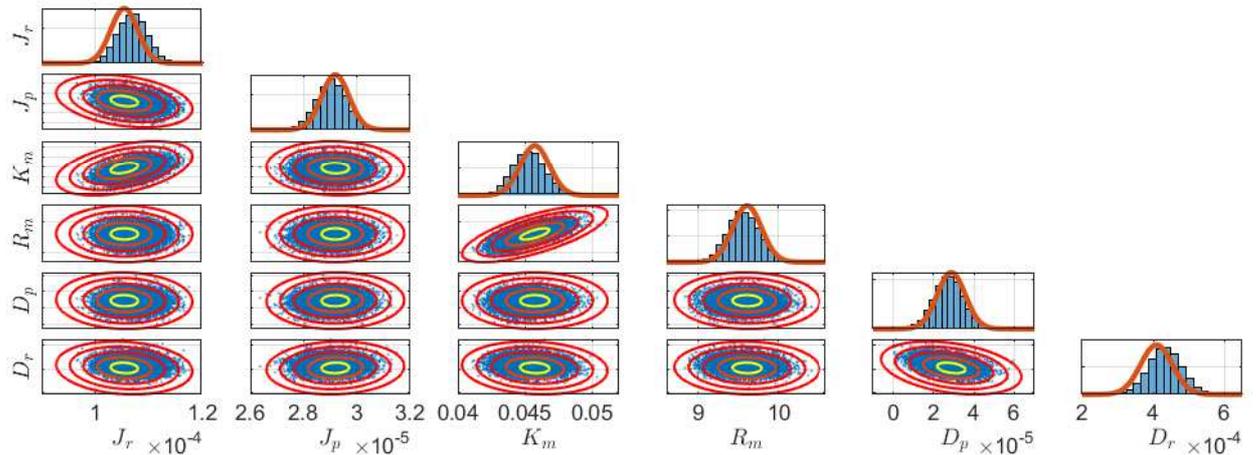}
 \caption{Distribution of inverted pendulum parameters calculated using the proposed method (contour plots) and HMC (scatter plots / histograms).}
 \label{fig:pend_params}
\end{figure*}

In Figure~\ref{fig:pend_params} the distribution of \(\phi\) using both the proposed approach and HMC are shown. This illustrates that, despite the approximations introduced, the proposed method has delivered results close to HMC. This has been achieved with a significant computational reduction. The proposed method required a few minutes to calculate while the HMC results required a few hours.

\section{Conclusion} \label{sec:conclusion}
In this paper, a variational based approach to estimating state and parameter distributions is presented for general nonlinear state-space models. The proposed method results in an optimisation problem of a standard form for which, due to the parametrisation selected, derivatives are readily available. While the proposed method assumes a Gaussian density, and is therefore limited compared to more general HMC approaches, the numerical examples illustrate that approximations suitable for many purposes are obtained on both simulated and real data. Compared to state-of-the-art HMC approaches this is achieved at a significantly lower computational cost.

Future research directions for this work include blending the proposed approach with sequential Monte Carlo (SMC) algorithms along the lines of what has been done in e.g. \citep{Lindsten2018,Naesseth2020}. The motivation for this is to combine the computational efficiency of the proposed method and the theoretical guarantees of SMC to overcome the limitations imposed by the Gaussian assumed densities.

\bibliographystyle{plain}       
\bibliography{root} 
                                                   
\appendix
\section{Decomposition of \(\mathcal{L}\left(\beta\right)\)} \label{sec:integral_decomposition}
In this section, the decomposition of \(\mathcal{L}\left(\beta\right)\) required to express \(\mathcal{L}\left(\beta\right)\) as pairwise joints distributions, rather than as a full distribution is presented. This utilises the Markovian nature of state-space models in a fashion similar to related system identification approaches such as \citep{Schoen2011,Chitralekha2009,Saerkkae2013,Vrettas2008,Courts2020}. 

Firstly, \(\mathcal{L}\left(\beta\right)\) is expressed in the form of two integrals as
 \begin{align*}
  \mathcal{L}\left(\beta\right) &= \int q_\beta(\xAll, \theta) \log p(\xAll, \theta, \yAll) d\xAllTheta \\
  &\quad- \int q_\beta(\xAll, \theta) \log q_\beta(\xAll,\theta) d\xAllTheta.
 \end{align*}

The first integral is decomposed according to
 \begin{align*}
  &\int q_\beta(\xAll, \theta) \log p(\xAll, \theta, \yAll) d\xAllTheta \\
  &\quad=\int q_\beta(\xAll, \theta) \log p(\xAll, \yAll \mid \theta) d\xAllTheta  +  \int q_\beta(\xAll, \theta) \log  p\left(\theta\right) d\xAllTheta \notag\\
  &\quad=\int q_\beta(\xAll, \theta) \log p(x_1 \mid \theta) d\xAllTheta \\
  &\quad\quad + \sum_{k=1}^{T}  \int q_\beta(\xAll, \theta) \log p(x_{k+1}, y_k \mid x_k, \theta) d\xAllTheta \notag\\
  &\quad\quad +  \int q_\beta(\xAll, \theta) \log  p\left(\theta\right) d\xAllTheta \notag\\
  &\quad=\int q_\beta(\xAll, \theta) \log p(x_1, \theta) d\xAllTheta \\
  &\quad\quad + \sum_{k=1}^{T}  \int q_\beta(\xAll, \theta) \log p(x_{k+1}, y_k \mid x_k, \theta) d\xAllTheta \notag \\
  &\quad=\int q_\beta(x_1, \theta) \log p(x_1, \theta) d[x_1, \theta] \\
  &\quad\quad + \sum_{k=1}^{T}  \int q_\beta(x_{k:k+1}, \theta) \log p(x_{k+1}, y_k \mid x_k, \theta) d[x_{k:k+1}, \theta] \notag\\
  &\quad=I_1\left(\beta\right) + I_{23}\left(\beta\right).
 \end{align*}

While the second integral is decomposed according to
\begin{align*}
	&\int q_\beta(\xAll, \theta) \log q_\beta(\xAll,\theta) d\xAllTheta \\
	&\quad= \int q_\beta(\xAll, \theta) \log q_\beta(\xAll \mid\theta) d\xAllTheta + \int q_\beta(\xAll, \theta) \log q_\beta(\theta) d\xAllTheta\notag\\
	&\quad=\int q_\beta(\xAll, \theta) \log q_\beta(x_1 \mid\theta) d\xAllTheta \\
	&\quad\quad+ \sum_{k=1}^{T}\int q_\beta(x_{\xAll}, \theta) \log q_\beta(x_{k+1} \mid x_k, \theta) d\xAllTheta \notag \\
	&\quad\quad+\int q_\beta(\xAll, \theta) \log q_\beta(\theta) d\xAllTheta \notag \\
	&\quad=\int q_\beta(\xAll, \theta) \log q_\beta(x_1, \theta) d\xAllTheta \\
	&\quad\quad+ \sum_{k=1}^{T}\int q_\beta(x_{\xAll}, \theta) \log \frac{q_\beta(x_{k+1}, x_k, \theta) }{q_\beta( x_k, \theta)} d\xAllTheta. \notag \\	
\end{align*}
As
\begin{align*}
	\log \frac{q_\beta(x_{k+1}, x_k, \theta) }{q_\beta( x_k, \theta)} =  \log q_\beta(x_{k+1}, x_k, \theta) - \log q_\beta(x_k, \theta),
\end{align*}
this gives
\begin{align*}
	&\int q_\beta(\xAll, \theta) \log q_\beta(\xAll,\theta) d\xAllTheta \\
	&\quad= \sum_{k=1}^{T}\int q_\beta(\xAll, \theta) \log q_\beta(x_{k+1}, x_k, \theta) d\xAllTheta \\
	&\quad\quad- \sum_{k=2}^{T}\int q_\beta(\xAll, \theta) \log q_\beta(x_k, \theta) d\xAllTheta, \notag \\
\end{align*}
and therefore,
\begin{align*}
	&\int q_\beta(\xAll, \theta) \log q_\beta(\xAll,\theta) d\xAllTheta \\
	&\quad= \sum_{k=1}^{T}\int q_\beta(x_{k+1}, x_k, \theta) \log q_\beta(x_{k+1}, x_k, \theta) d[x_{k:k+1}, \theta] \notag\\
	&\quad\quad- \sum_{k=2}^{T}\int q_\beta( x_k, \theta) \log q_\beta(x_k, \theta) d[x_{k}, \theta]  \\
	&\quad= I_4\left(\beta\right).
\end{align*}

Together, this allows \(\mathcal{L}\left(\beta\right)\) to be calculated without requiring the full distribution, \( q_\beta(\xAll, \theta) \), but rather only each \(q_\beta(\theta, x_k, x_{k+1}) \) using
\begin{align*}
 \mathcal{L}\left(\beta\right) = I_1\left(\beta\right) + I_{23}\left(\beta\right) - I_4\left(\beta\right).
\end{align*}

\section{Inverted Pendulum Model} \label{sec:pendulum_model}
 The state vector used to model the Furata pendulum is
\(
x = \begin{bmatrix}
 \vartheta &
 \alpha &
 \dot{\vartheta} &
 \dot{\alpha}
\end{bmatrix}^\Transp, 
\)
where \(\vartheta\) and \(\alpha\) are the arm and pendulum angles, respectively. 

The continuous time dynamics are given by,
 \begin{align*}
  \dot{x} = \begin{bmatrix}
   \dot{\vartheta} &
   \dot{\alpha} &
   \ddot{\vartheta} &
   \ddot{\alpha}
  \end{bmatrix}^\Transp,
 \end{align*}
 where
 \begin{align*}
  \begin{bmatrix}
   \ddot{\vartheta} \\
   \ddot{\alpha}
  \end{bmatrix} = M^{-1}\left(\begin{bmatrix}
   \tau - D_r \dot{\vartheta}\\
   -D_p \dot{\alpha} - \frac{1}{2}m_p l_p g \sin\left(\alpha\right)
  \end{bmatrix} - C\right),
 \end{align*}
 and
 \begin{align*}
  C &= \begin{bmatrix}
   \frac{1}{2}m_p l_p^2 \sin\left(\alpha\right) \cos\left(\alpha\right) \dot{\vartheta} \dot{\alpha} - \frac{1}{2}m_p l_p l_r \sin\left(\alpha\right) \dot{\alpha}^2 \\
   -\frac{1}{4} m_p l_p^2 \cos\left(\alpha\right) \sin\left(\alpha\right) \dot{\vartheta}^2
  \end{bmatrix}, \\
  \tau &= \frac{k_m\left(V_m - k_m \dot{\vartheta}\right)}{R_m}, \\
  M &= \begin{bmatrix}
   M_{11} & M_{12} \\
   M_{21} & M_{22}
  \end{bmatrix},
 \end{align*}
 where
 \begin{align*}
  M_{11} &= J_r + m_p l_r^2 + \frac{1}{4}\left( m_pl_p^2 - m_p l_p^2 \cos^2\left(\alpha\right)\right), \\
  M_{12} &= M_{21} \\
  &= \frac{1}{2}m_p l_p l_r \cos^2\left(\alpha\right), \\
  M_{22} &= J_p + \frac{1}{4}m_p l_p^2,
 \end{align*}
and \(m_p\) is the pendulum mass, \(l_r\), \(l_p\) and the rod and pendulum lengths respectively, \(J_r\), \(J_p\) are the rod and pendulum inertias, \(R_m\) is the motor resistance, \(k_m\) is the motor constant, \(D_r\), \(D_p\) are damping coefficients for the rod and pendulum respectively, and \(V_m\) is the applied motor voltage input. 

The process model used consists of a two-step Euler discretisation of these continuous time dynamics over an \SI{8}{\milli\second} sampling time subsequently disturbed by noise \(v_k\).

Measurements are from encoders on the arm and pendulum angle and current measurement from the motor. The resultant measurement model is
\begin{align*}
 y_k &= \begin{bmatrix}
  \vartheta &
  \alpha &
  \frac{V_m - k_m \dot{\vartheta}}{R_m}
 \end{bmatrix}^\Transp + w_k,
\end{align*}
and it is assumed that
\begin{align*}
 \begin{bmatrix}
  v_k \\
  w_k
 \end{bmatrix} \sim \mathcal{N}\left(0,\Pi\right).
\end{align*}

\end{document}